%% file: iclr2023_conference_tinypaper.tex
\documentclass{article} 
\usepackage{iclr2023_conference_tinypaper,times}

\input{math_commands.tex}

\usepackage{hyperref}
\usepackage{url}
\usepackage{graphicx}
\usepackage{tabularx}
\usepackage{longtable}
\usepackage{subcaption}

\title{cognitive resilience: unraveling the proficiency of image-captioning models to interpret masked visual content}

\author{Zhicheng Du$^1$, Zhaotian Xie$^1$, Huazhang Ying$^1$, Likun Zhang$^1$, Peiwu Qin$^{1,}\thanks{Corresponding author}$ \\
$^1$Tsinghua Shenzhen International Graduate School, Tsinghua University \\
\texttt{\{duzc21, xzt21, yhz22, zlk21\}@mails.tsinghua.edu.cn}\\ 
\texttt{pwqin@sz.tsinghua.edu.cn}\\ 
}

\iclrfinalcopy 
\begin{document}

\maketitle
\begin{abstract}
This study explores the ability of Image Captioning (IC) models to decode masked visual content sourced from diverse datasets. Our findings reveal the IC model's capability to generate captions from masked images, closely resembling the original content. Notably, even in the presence of masks, the model adeptly crafts descriptive textual information that goes beyond what is observable in the original image-generated captions. While the decoding performance of the IC model experiences a decline with an increase in the masked region's area, the model still performs well when important regions of the image are not masked at high coverage.
\end{abstract}

\section{Introduction}
\label{int}
Image Captioning (IC), also known as the Image-to-Text task, is a pivotal multimodal research area whose goal is to generate natural language descriptions of visual content \citep{DBLP:journals/air/SharmaP23, DBLP:conf/hci/Ding23}. The IC model aims to furnish a holistic comprehension of the image, capturing object relationships, attributes, scene characteristics, and interactions among objects within the given scene \citep{imgc}. Recent years have borne witness to a pronounced proliferation of scholarly pursuits in the sphere of IC, culminating in a manifold of applications spanning diverse domains, including biomedical IC \citep{DBLP:conf/icml/Zheng023} and bespoke assistive technology catering to the visually impaired \citep{DBLP:conf/bibm/GuoCXBO22}. Concomitantly, the demonstrated efficacy of the masked autoencoder (MAE) \citep{DBLP:conf/cvpr/HeCXLDG22} highlights the potential latent in visual self-supervised learning (SSL), which has attracted considerable attention in particular for its aptitude in predicting masked input attributes predicated upon unmasked input content \citep{DBLP:journals/corr/abs-2208-00173}. Inspired by the success of MAE, we raise an incisive inquiry: Can IC models directly discern the masked visual content and yield accurate textual descriptions? While existing IC models have demonstrated proficiency in generating precise captions for clear images, limited insights exist into their capacity to comprehend masked visual content. Herein, we design several masking methodologies applied to the image, aiming to scrutinize the impact of diverse masking strategies on the captions generated by the model.

\section{Method}
\label{met}
The overall pipeline of this work is shown in Appendix \ref{sa}. We assess four IC models, including the Large Language and Vision Assistant (LLaVA) \citep{liu2023improved}, ViT-GPT2-Image-Captioning model \citep{kumar2022imagecaptioning}, Generative Image-to-Text Transformer (GIT) \citep{DBLP:journals/tmlr/WangYHLLGLLW22}, and Bootstrapping Language-Image Pre-Training (BLIP) \citep{DBLP:conf/icml/0001LXH22}. We design three distinct masking methods (i.e., masked ratio, masked block size, and color) for image masking. The masked ratio represents the proportion of the covered image area to the total image size; masked block size denotes the area of the square used for random masking; and color means the hue of the masked block. A random selection of masked areas combines these ways to mask the image. We conduct both quantitative and qualitative analyses to scrutinize the disparities in textual descriptions generated from the original and masked images. The pre-trained General Text Embeddings (GTE) model \citep{li2023general} is used to compute semantic textual similarity scores as a criterion for quantitative evaluation.

\section{Experiments}
\label{exp}

\textbf{Dataset and Experimental setting.} We construct a dataset comprising 60 images to deploy the performance evaluation of the IC model. To minimize the impact of chance and ensure a diverse representation, we randomly select ten images from each of the six distinct source datasets (i.e., the Pokemon\footnote{https://huggingface.co/datasets/diffusers/pokemon-gpt4-captions}, COCO2017 \citep{DBLP:conf/eccv/LinMBHPRDZ14}, LAION-COCO \citep{DBLP:conf/nips/SchuhmannBVGWCC22}, Impressions \citep{kruk2023impressions}, Midjourney-threads\citep{DBLP:journals/corr/abs-2311-12131}, and Flickr8k \citep{flickr} datasets). The distribution of these images is provided in Appendix \ref{sb}. The masked ratio incrementally ascends in 5\% intervals, while the square masked block's side length takes on values of 4, 8, 16, 64, 128, and 256. The color of the masked block includes black, white, and grey.

\begin{figure}[ht]
\includegraphics[width=1\linewidth]{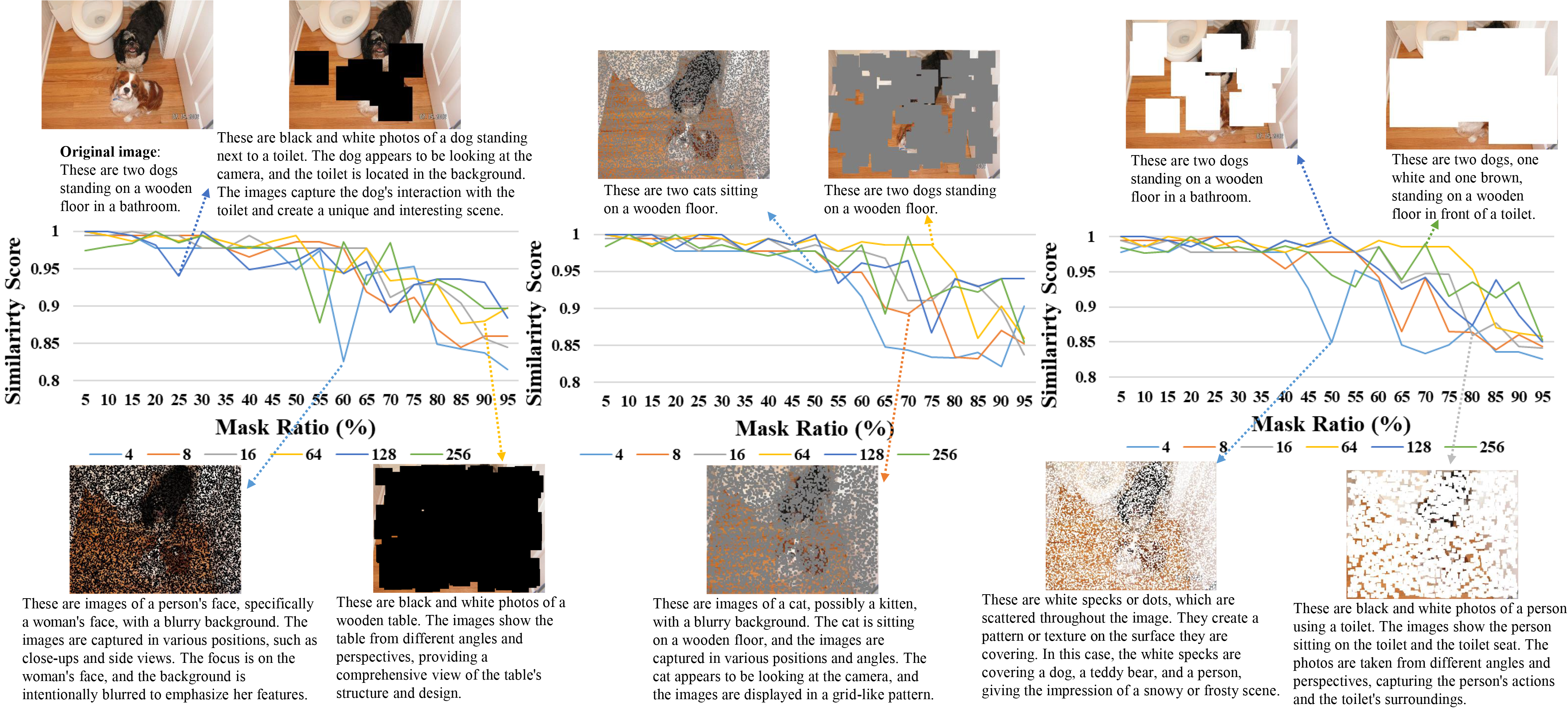}
\caption{The quantitative and qualitative outcome results of LLaVA. Three line charts represent the colors of masked block, where the left one is black, middle one is gray, and the right one is white.}
\label{fig1}
\end{figure}

\textbf{Results}. Figure \ref{fig1} illustrates the quantitative and qualitative outcomes of the LLaVA. The upper and lower sections of the image showcase examples of qualitative analysis, depicting high and low semantic similarity between the captions of the masked image and the subtitles of the original image, respectively. In the middle, we observe the quantitative change in semantic similarity as the masked ratio increases. Our findings reveal a non-linear correlation between the captions generated by the IC model and the original image as the random masking rate increases. Surprisingly, the accuracy of captions corresponding to images with a low masking rate is higher than that of those with a high masking rate. Our study elucidates that the model exhibits heightened accuracy in caption generation when pivotal and contextually relevant information within the image remains unmasked, contingent upon the positioning of the masking block. Notably, larger masking blocks in most instances increase the likelihood of concealing pertinent information within the image region, leading the model to produce inaccurate descriptions even when the masking rate is low. Interestingly, we observe that the color of the masked block can erroneously influence the IC model to output information not present in the image (see Appendix \ref{sc}). Moreover, the masking process induces the IC model to output information absent in the original caption (see Appendix \ref{sd}). These insights highlight the nuanced impact of masking strategies on the IC model's output.

\section{Discussion and Conclusion}
\label{con}
Our study comprehensively analyzes the ability of IC models to understand masked visual content across various conditions and degrees. The relevance of three distinct approaches to mask processing and the subsequent elaboration of textual descriptions generated by IC models is highlighted. In future research endeavors, we plan to delve deeper into mining the relationships and importance ranking between different regions within an image and exploring innovative image masking processing methods to further advance the development of visual SSL and multimodal models.

\subsubsection*{URM Statement}
The authors acknowledge that all key authors of this work meet the URM criteria of ICLR 2024 Tiny Papers Track.

\subsubsection*{Reproducibility Statement}
The source code, dataset used in this study, and experimental results will be available at https://github.com/dodoxxb/cognitive-resilience.

\subsubsection*{Acknowledgements}
This research is supported by the National Natural Science Foundation of China 31970752,32350410397;Science,Technology,Innovation Commission of Shenzhen Municipality,JCYJ20220530143014032,JCYJ20230807113017035,WDZC20200820173710001,Shenzhen Medical Research Funds,D2301002;Department of Chemical Engineering-iBHE special cooperation joint fund project, DCE-iBHE-2022-3; Tsinghua Shenzhen Interna-tional Graduate School Cross-disciplinary Research and Innovation Fund Research Plan, JC2022009; and Bureau of Planning, Land and Resources of Shenzhen Municipality (2022) 207.

\bibliography{iclr2023_conference_tinypaper}
\bibliographystyle{iclr2023_conference_tinypaper}

\clearpage
\appendix
\section{The overall pipeline of this study}
\label{sa}
\begin{figure}[h]
\includegraphics[width=1\linewidth]{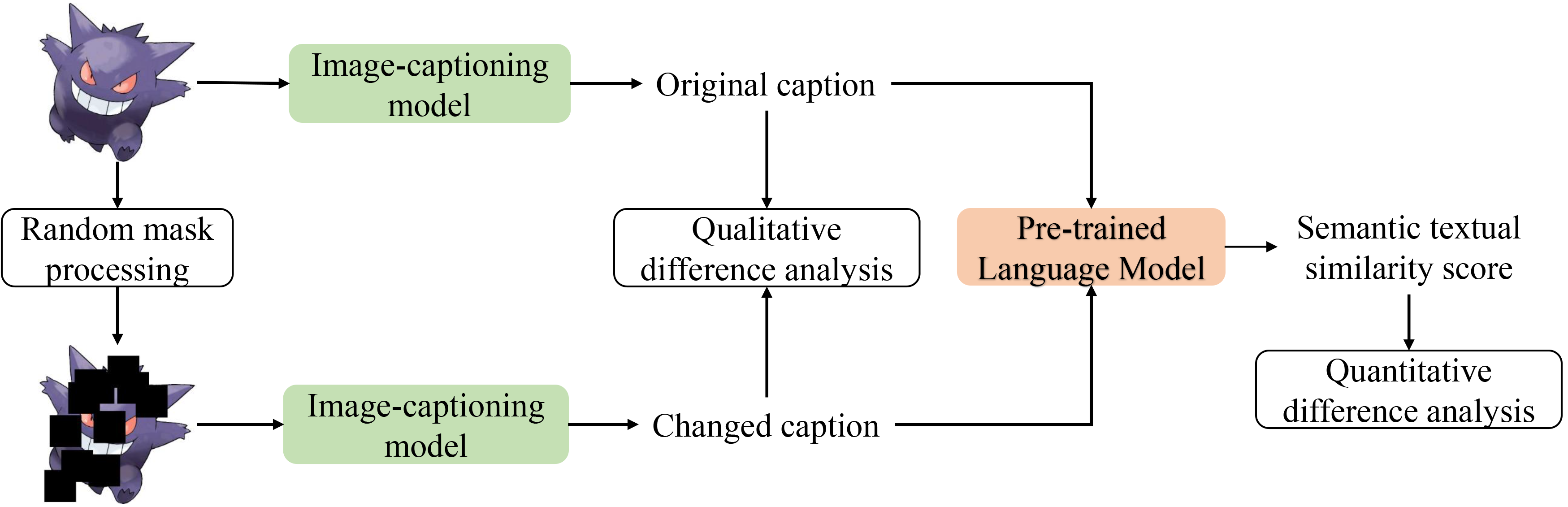}
\caption{Overall pipeline of this work, where the random mask processing includes three ways (i.e., masked ratio, masked block size, and color), and the pre-trained GTE model are used to deploy the experiment of semantic textual similarity.}
\label{fig2}
\end{figure}

\section{The t-distributed stochastic neighbor embedding (t-SNE) visualization of the dataset}
\label{sb}
We employ the ResNet-50 model \citep{DBLP:conf/cvpr/HeZRS16} to extract features from the dataset used in this study. Subsequently, we utilize the t-SNE algorithm \citep{JMLR:v9:vandermaaten08a} to visualize the distribution of the dataset.
\begin{figure}[ht]
\includegraphics[width=1\linewidth]{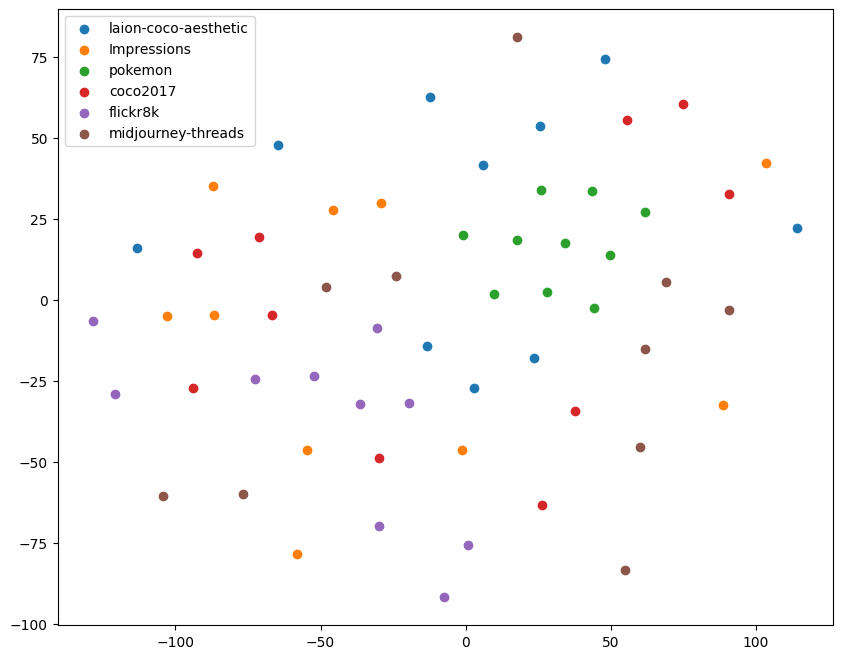}
\caption{The distribution of the dataset used in this work}
\label{fig3}
\end{figure}  

\section{The effect of the masked block's color}
\label{sc}
The qualitative analysis demonstrates that incorporating the masked operation with diverse colors has a noticeable impact on the IC model's output, resulting in more and varied information under equivalent conditions. The strategic use of different colors proves advantageous, enriching the model's ability to provide a more detailed image description. Even in instances where the IC model encounters difficulties in accurately portraying the original image, the meticulous application of the masking process can play a crucial role in addressing these challenges and facilitating improvements.
\begin{figure}[ht]
\includegraphics[width=1\linewidth]{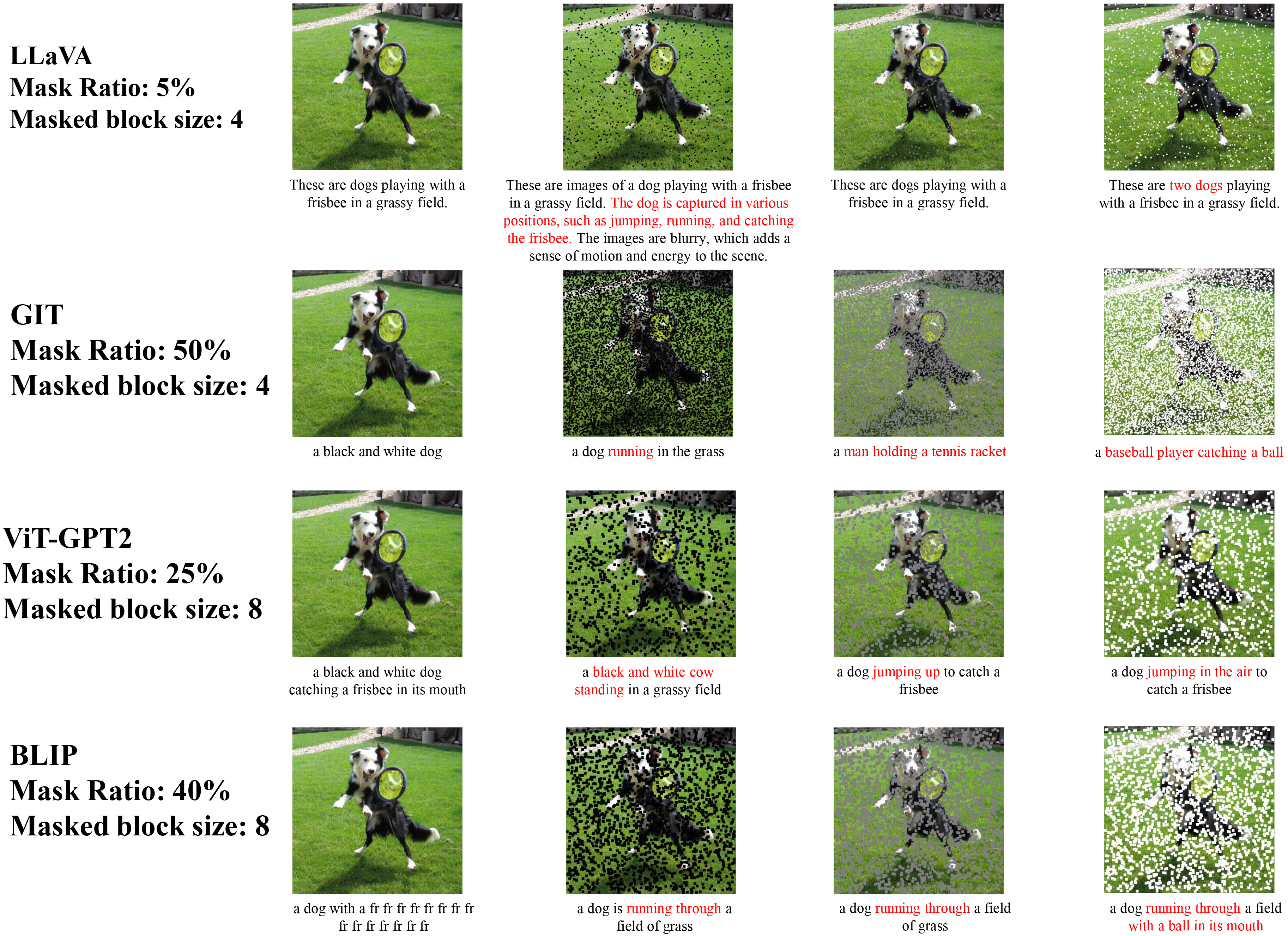}
\caption{The results of the interpretation of images processed with different mask block colors, where the text in red means the differences with the text from the original image. The first column is the original image, and the second through fourth columns are the images processed with black, gray and white masks respectively (image example from the flickr8k dataset and the used IC model is LLaVA).}
\label{fig4}
\end{figure}

\clearpage
\section{The supplementary effect of the mask processing for the IC model}
\label{sd}
Through qualitative analysis, we observe that a thoughtfully designed masked operation enhances the ability of the IC model to capture additional hidden information in images. It's worth noting that this newfound information may occasionally be incorrect, representing the model's speculative understanding of the image. Nevertheless, the positive aspect is that the masked operation enables the IC model to delve into more nuanced details of the image, showcasing an improved understanding.
\begin{figure}[ht]
\includegraphics[width=1\linewidth]{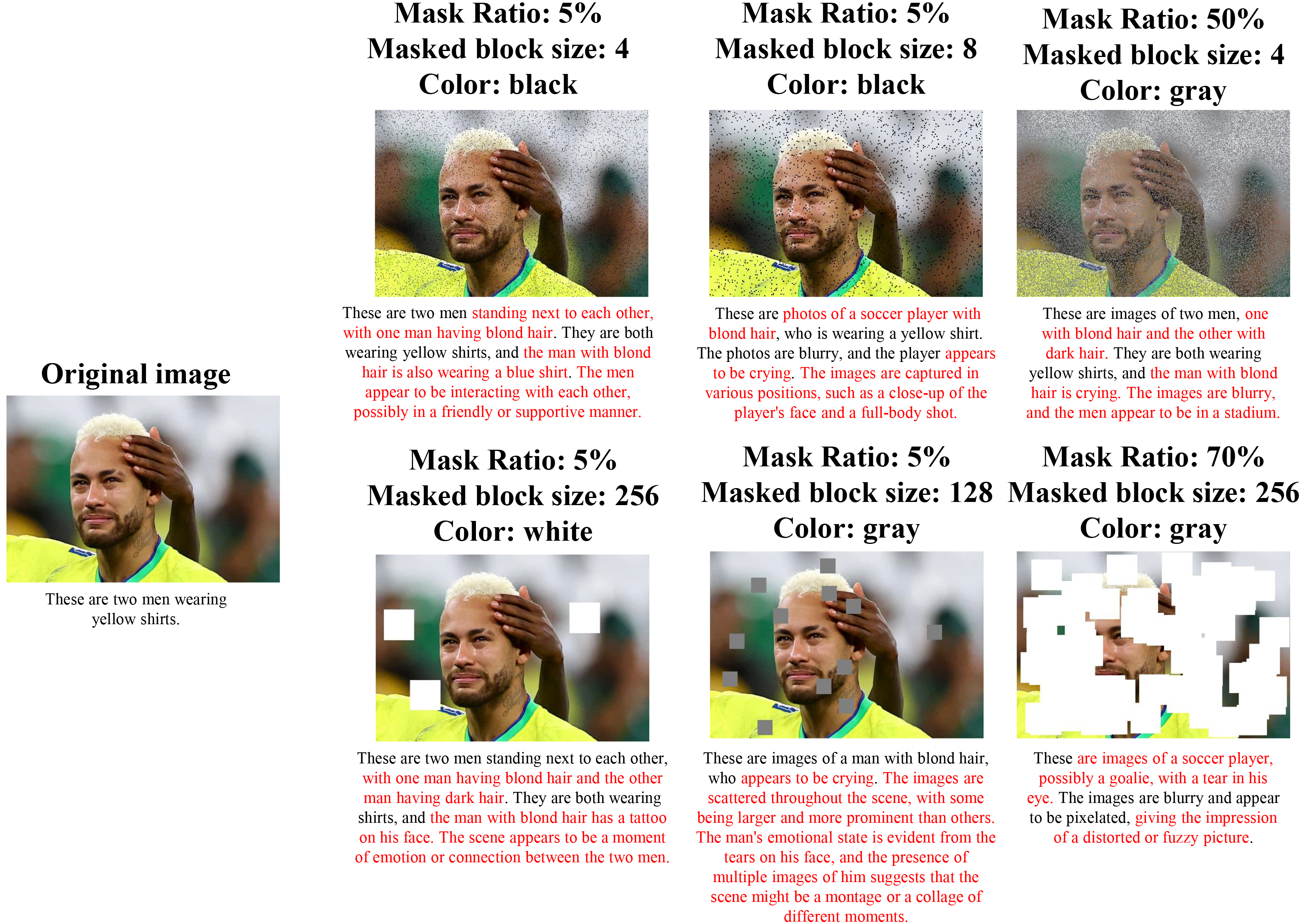}
\caption{The results of showing mask processing help the IC model output more informative results, where the text in red means it supplements the text from the original image (image example from the flickr8k dataset and the used IC model is LLaVA). }
\label{fig5}
\end{figure}

\end{document}

%% file: math_commands.tex
\usepackage{amsmath,amsfonts,bm}

\def\eqref#1{equation~\ref{#1}}

\def\1{\bm{1}}

\DeclareMathAlphabet{\mathsfit}{\encodingdefault}{\sfdefault}{m}{sl}
\SetMathAlphabet{\mathsfit}{bold}{\encodingdefault}{\sfdefault}{bx}{n}













%% file: iclr2023_conference_tinypaper.bbl
\begin{thebibliography}{19}
\providecommand{\natexlab}[1]{#1}
\providecommand{\url}[1]{\texttt{#1}}
\expandafter\ifx\csname urlstyle\endcsname\relax
  \providecommand{\doi}[1]{doi: #1}\else
  \providecommand{\doi}{doi: \begingroup \urlstyle{rm}\Url}\fi

\bibitem[Ding(2023)]{DBLP:conf/hci/Ding23}
Yi~Ding.
\newblock A systematic literature review on image captioning.
\newblock In Constantine Stephanidis, Margherita Antona, Stavroula Ntoa, and Gavriel Salvendy (eds.), \emph{{HCI} International 2023 Posters - 25th International Conference on Human-Computer Interaction, {HCII} 2023, Copenhagen, Denmark, July 23-28, 2023, Proceedings, Part {V}}, volume 1836 of \emph{Communications in Computer and Information Science}, pp.\  396--404. Springer, 2023.
\newblock \doi{10.1007/978-3-031-36004-6\_54}.
\newblock URL \url{https://doi.org/10.1007/978-3-031-36004-6\_54}.

\bibitem[Don{-}Yehiya et~al.(2023)Don{-}Yehiya, Choshen, and Abend]{DBLP:journals/corr/abs-2311-12131}
Shachar Don{-}Yehiya, Leshem Choshen, and Omri Abend.
\newblock Human learning by model feedback: The dynamics of iterative prompting with midjourney.
\newblock \emph{CoRR}, abs/2311.12131, 2023.
\newblock \doi{10.48550/ARXIV.2311.12131}.
\newblock URL \url{https://doi.org/10.48550/arXiv.2311.12131}.

\bibitem[Ghandi et~al.(2023)Ghandi, Pourreza, and Mahyar]{imgc}
Taraneh Ghandi, Hamidreza Pourreza, and Hamidreza Mahyar.
\newblock Deep learning approaches on image captioning: A review.
\newblock \emph{ACM Comput. Surv.}, 56\penalty0 (3), oct 2023.
\newblock ISSN 0360-0300.
\newblock \doi{10.1145/3617592}.
\newblock URL \url{https://doi.org/10.1145/3617592}.

\bibitem[Guo et~al.(2022)Guo, Chen, Xie, Ban, and Obaidat]{DBLP:conf/bibm/GuoCXBO22}
Yu~Guo, Yue Chen, Yuanyan Xie, Xiaojuan Ban, and Mohammad~S. Obaidat.
\newblock An offline assistance tool for visually impaired people based on image captioning.
\newblock In Donald~A. Adjeroh, Qi~Long, Xinghua~Mindy Shi, Fei Guo, Xiaohua Hu, Srinivas Aluru, Giri Narasimhan, Jianxin Wang, Mingon Kang, Ananda Mondal, and Jin Liu (eds.), \emph{{IEEE} International Conference on Bioinformatics and Biomedicine, {BIBM} 2022, Las Vegas, NV, USA, December 6-8, 2022}, pp.\  969--976. {IEEE}, 2022.
\newblock \doi{10.1109/BIBM55620.2022.9994947}.
\newblock URL \url{https://doi.org/10.1109/BIBM55620.2022.9994947}.

\bibitem[He et~al.(2016)He, Zhang, Ren, and Sun]{DBLP:conf/cvpr/HeZRS16}
Kaiming He, Xiangyu Zhang, Shaoqing Ren, and Jian Sun.
\newblock Deep residual learning for image recognition.
\newblock In \emph{2016 {IEEE} Conference on Computer Vision and Pattern Recognition, {CVPR} 2016, Las Vegas, NV, USA, June 27-30, 2016}, pp.\  770--778. {IEEE} Computer Society, 2016.
\newblock \doi{10.1109/CVPR.2016.90}.
\newblock URL \url{https://doi.org/10.1109/CVPR.2016.90}.

\bibitem[He et~al.(2022)He, Chen, Xie, Li, Doll{\'{a}}r, and Girshick]{DBLP:conf/cvpr/HeCXLDG22}
Kaiming He, Xinlei Chen, Saining Xie, Yanghao Li, Piotr Doll{\'{a}}r, and Ross~B. Girshick.
\newblock Masked autoencoders are scalable vision learners.
\newblock In \emph{{IEEE/CVF} Conference on Computer Vision and Pattern Recognition, {CVPR} 2022, New Orleans, LA, USA, June 18-24, 2022}, pp.\  15979--15988. {IEEE}, 2022.
\newblock \doi{10.1109/CVPR52688.2022.01553}.
\newblock URL \url{https://doi.org/10.1109/CVPR52688.2022.01553}.

\bibitem[Hodosh et~al.(2013)Hodosh, Young, and Hockenmaier]{flickr}
Micah Hodosh, Peter Young, and Julia Hockenmaier.
\newblock Framing image description as a ranking task: Data, models and evaluation metrics.
\newblock \emph{J. Artif. Int. Res.}, 47\penalty0 (1):\penalty0 853–899, may 2013.
\newblock ISSN 1076-9757.

\bibitem[Kruk et~al.(2023)Kruk, Ziems, and Yang]{kruk2023impressions}
Julia Kruk, Caleb Ziems, and Diyi Yang.
\newblock Impressions: Understanding visual semiotics and aesthetic impact, 2023.

\bibitem[Kumar(2022)]{kumar2022imagecaptioning}
Ankur Kumar.
\newblock The illustrated image captioning using transformers.
\newblock \emph{ankur3107.github.io}, 2022.
\newblock URL \url{https://ankur3107.github.io/blogs/the-illustrated-image-captioning-using-transformers/}.

\bibitem[Li et~al.(2022)Li, Li, Xiong, and Hoi]{DBLP:conf/icml/0001LXH22}
Junnan Li, Dongxu Li, Caiming Xiong, and Steven C.~H. Hoi.
\newblock {BLIP:} bootstrapping language-image pre-training for unified vision-language understanding and generation.
\newblock In Kamalika Chaudhuri, Stefanie Jegelka, Le~Song, Csaba Szepesv{\'{a}}ri, Gang Niu, and Sivan Sabato (eds.), \emph{International Conference on Machine Learning, {ICML} 2022, 17-23 July 2022, Baltimore, Maryland, {USA}}, volume 162 of \emph{Proceedings of Machine Learning Research}, pp.\  12888--12900. {PMLR}, 2022.
\newblock URL \url{https://proceedings.mlr.press/v162/li22n.html}.

\bibitem[Li et~al.(2023)Li, Zhang, Zhang, Long, Xie, and Zhang]{li2023general}
Zehan Li, Xin Zhang, Yanzhao Zhang, Dingkun Long, Pengjun Xie, and Meishan Zhang.
\newblock Towards general text embeddings with multi-stage contrastive learning, 2023.

\bibitem[Lin et~al.(2014)Lin, Maire, Belongie, Hays, Perona, Ramanan, Doll{\'{a}}r, and Zitnick]{DBLP:conf/eccv/LinMBHPRDZ14}
Tsung{-}Yi Lin, Michael Maire, Serge~J. Belongie, James Hays, Pietro Perona, Deva Ramanan, Piotr Doll{\'{a}}r, and C.~Lawrence Zitnick.
\newblock Microsoft {COCO:} common objects in context.
\newblock In David~J. Fleet, Tom{\'{a}}s Pajdla, Bernt Schiele, and Tinne Tuytelaars (eds.), \emph{Computer Vision - {ECCV} 2014 - 13th European Conference, Zurich, Switzerland, September 6-12, 2014, Proceedings, Part {V}}, volume 8693 of \emph{Lecture Notes in Computer Science}, pp.\  740--755. Springer, 2014.
\newblock \doi{10.1007/978-3-319-10602-1\_48}.
\newblock URL \url{https://doi.org/10.1007/978-3-319-10602-1\_48}.

\bibitem[Liu et~al.(2023)Liu, Li, Li, and Lee]{liu2023improved}
Haotian Liu, Chunyuan Li, Yuheng Li, and Yong~Jae Lee.
\newblock Improved baselines with visual instruction tuning, 2023.

\bibitem[Schuhmann et~al.(2022)Schuhmann, Beaumont, Vencu, Gordon, Wightman, Cherti, Coombes, Katta, Mullis, Wortsman, Schramowski, Kundurthy, Crowson, Schmidt, Kaczmarczyk, and Jitsev]{DBLP:conf/nips/SchuhmannBVGWCC22}
Christoph Schuhmann, Romain Beaumont, Richard Vencu, Cade Gordon, Ross Wightman, Mehdi Cherti, Theo Coombes, Aarush Katta, Clayton Mullis, Mitchell Wortsman, Patrick Schramowski, Srivatsa Kundurthy, Katherine Crowson, Ludwig Schmidt, Robert Kaczmarczyk, and Jenia Jitsev.
\newblock {LAION-5B:} an open large-scale dataset for training next generation image-text models.
\newblock In \emph{NeurIPS}, 2022.
\newblock URL \url{http://papers.nips.cc/paper\_files/paper/2022/hash/a1859debfb3b59d094f3504d5ebb6c25-Abstract-Datasets\_and\_Benchmarks.html}.

\bibitem[Sharma \& Padha(2023)Sharma and Padha]{DBLP:journals/air/SharmaP23}
Himanshu Sharma and Devanand Padha.
\newblock A comprehensive survey on image captioning: from handcrafted to deep learning-based techniques, a taxonomy and open research issues.
\newblock \emph{Artif. Intell. Rev.}, 56\penalty0 (11):\penalty0 13619--13661, 2023.
\newblock \doi{10.1007/S10462-023-10488-2}.
\newblock URL \url{https://doi.org/10.1007/s10462-023-10488-2}.

\bibitem[van~der Maaten \& Hinton(2008)van~der Maaten and Hinton]{JMLR:v9:vandermaaten08a}
Laurens van~der Maaten and Geoffrey Hinton.
\newblock Visualizing data using t-sne.
\newblock \emph{Journal of Machine Learning Research}, 9\penalty0 (86):\penalty0 2579--2605, 2008.
\newblock URL \url{http://jmlr.org/papers/v9/vandermaaten08a.html}.

\bibitem[Wang et~al.(2022)Wang, Yang, Hu, Li, Lin, Gan, Liu, Liu, and Wang]{DBLP:journals/tmlr/WangYHLLGLLW22}
Jianfeng Wang, Zhengyuan Yang, Xiaowei Hu, Linjie Li, Kevin Lin, Zhe Gan, Zicheng Liu, Ce~Liu, and Lijuan Wang.
\newblock {GIT:} {A} generative image-to-text transformer for vision and language.
\newblock \emph{Trans. Mach. Learn. Res.}, 2022, 2022.
\newblock URL \url{https://openreview.net/forum?id=b4tMhpN0JC}.

\bibitem[Zhang et~al.(2022)Zhang, Zhang, Song, Yi, Zhang, and Kweon]{DBLP:journals/corr/abs-2208-00173}
Chaoning Zhang, Chenshuang Zhang, Junha Song, John Seon~Keun Yi, Kang Zhang, and In~So Kweon.
\newblock A survey on masked autoencoder for self-supervised learning in vision and beyond.
\newblock \emph{CoRR}, abs/2208.00173, 2022.
\newblock \doi{10.48550/ARXIV.2208.00173}.
\newblock URL \url{https://doi.org/10.48550/arXiv.2208.00173}.

\bibitem[Zheng \& Yu(2023)Zheng and Yu]{DBLP:conf/icml/Zheng023}
Ervine Zheng and Qi~Yu.
\newblock Evidential interactive learning for medical image captioning.
\newblock In Andreas Krause, Emma Brunskill, Kyunghyun Cho, Barbara Engelhardt, Sivan Sabato, and Jonathan Scarlett (eds.), \emph{International Conference on Machine Learning, {ICML} 2023, 23-29 July 2023, Honolulu, Hawaii, {USA}}, volume 202 of \emph{Proceedings of Machine Learning Research}, pp.\  42478--42491. {PMLR}, 2023.
\newblock URL \url{https://proceedings.mlr.press/v202/zheng23g.html}.

\end{thebibliography}
